\documentclass[10pt,twocolumn,letterpaper]{article}

\usepackage{cvpr}              
\usepackage{bm}
\usepackage{multirow} 
\usepackage[table]{xcolor}

\definecolor{cvprblue}{rgb}{0.21,0.49,0.74}
\usepackage[pagebackref,breaklinks,colorlinks,allcolors=cvprblue]{hyperref}

\title{Towards High-Consistency Embodied World Model \\ with Multi-View Trajectory Videos}

\author{Taiyi Su$^1$, Jian Zhu$^1$\thanks{Corresponding Author~(jianzhu823@gmail.com)}, Yaxuan Li$^3$, Chong Ma$^2$, Jianjun Zhang$^2$, Zitai Huang$^2$, Hanli Wang$^2$, Yi Xu$^1$\\
$^1$Midea Group\\
$^2$Tongji University\\
$^3$East China Normal University\\
}

\begin{document}

\maketitle

\begin{abstract}
Embodied world models aim to predict and interact with the physical world through visual observations and actions. However, existing models struggle to accurately translate low-level actions (\textit{e.g.}, joint positions) into precise robotic movements in predicted frames, leading to inconsistencies with real-world physical interactions. To address these limitations, we propose MTV-World, an embodied world model that introduces \textbf{M}ulti-view \textbf{T}rajectory-\textbf{V}ideo control for precise visuomotor prediction. Specifically, instead of directly using low-level actions for control, we employ trajectory videos obtained through camera intrinsic and extrinsic parameters and Cartesian-space transformations as control signals. However, projecting 3D raw actions onto 2D images inevitably causes a loss of spatial information, making a single view insufficient for accurate interaction modeling. To overcome this, we introduce a multi-view framework that compensates for spatial information loss and ensures high consistency with the physical world. MTV-World forecasts future frames based on multi-view trajectory videos as input and conditioned on an initial frame per view. Furthermore, to systematically evaluate both robotic motion precision and object interaction accuracy, we develop an auto-evaluation pipeline leveraging multimodal large models and referring video object segmentation models. To measure spatial consistency, we formulate it as an object location matching problem and adopt the Jaccard Index as the evaluation metric. Extensive experiments demonstrate that MTV-World achieves precise control execution and accurate physical interaction modeling in complex dual-arm scenarios.
\end{abstract}

\section{Introduction}

\begin{figure}[htbp]
\centering
\includegraphics[width=\linewidth]{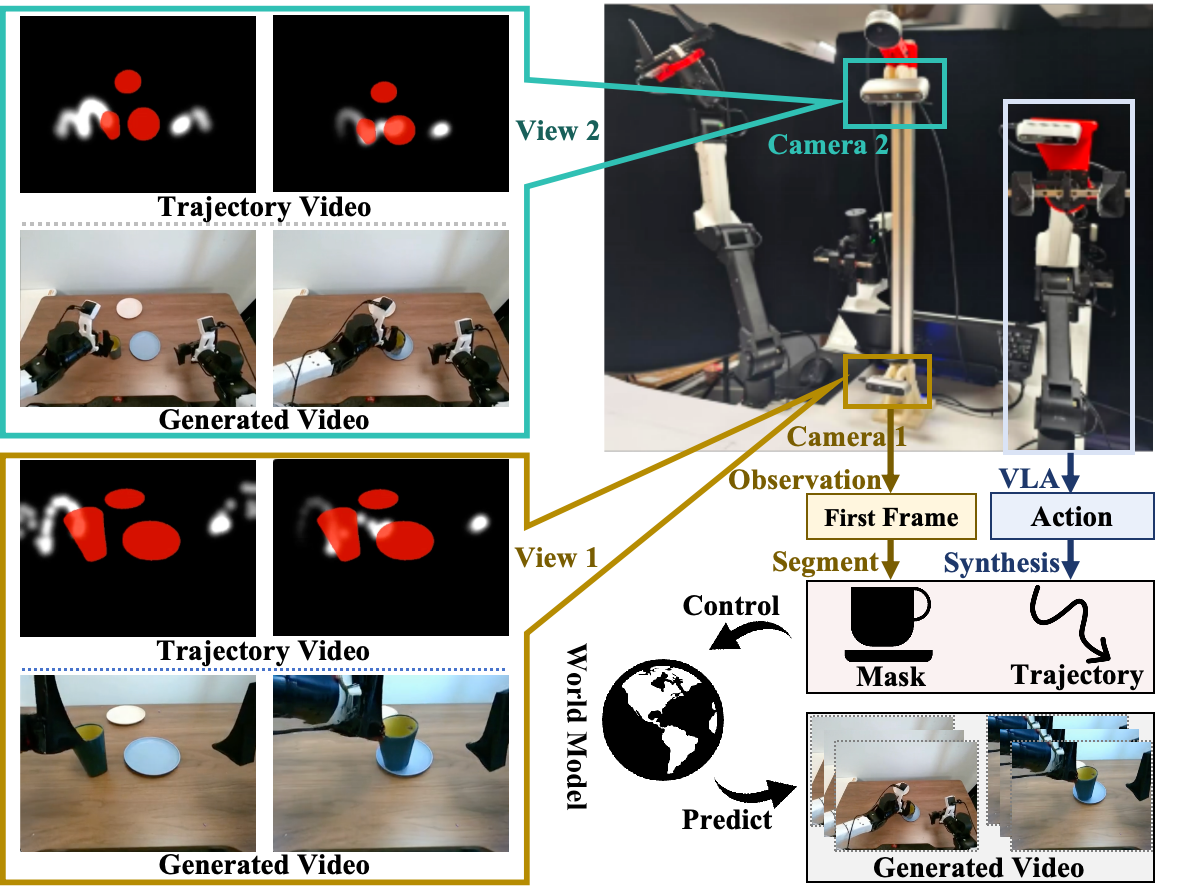}
\caption{
Overview of the proposed MTV-World, which utilizes multi-view trajectory videos as control inputs. Each trajectory video is synthesized by combining the initial object mask with trajectory motions for future frame prediction. The multi-view trajectories clearly depict the robot arm’s movement paths and its interactions with objects.}
\label{fig1}
\end{figure}

Recent advances in embodied intelligence have underscored the critical role of world models in enabling models to understand, predict, and interact with their environments~\cite{zhang2025matrix,hafner2025mastering,chi2025wow,bruce2024genie}. An embodied world model serves as a predictive simulator of the physical world, capable of forecasting future outcomes conditioned on current observations and planned actions~\cite{agarwal2025cosmos,guo2025ctrl,zhu2025unified}. In practical applications, an embodied world model can function as a data engine~\cite{jiang2025enerverse,huang2025enerverse}, generating diverse and controllable visual experiences to enhance policy learning, and as a policy evaluator~\cite{li2025worldeval,li2024evaluating,quevedo2025evaluating,zhou2025autoeval}, providing a safe and efficient environment to assess and refine decision-making strategies without real-world execution.

Despite rapid progress in embodied intelligence, current embodied world models still face significant limitations. Most existing models struggle to accurately translate low-level actions (\textit{e.g.}, joint positions) into precise robotic motions, resulting in unstable or unrealistic arm trajectories in predicted videos. Consequently, a second challenge naturally arises: when the predicted arm motion itself is imprecise, the model cannot reliably simulate object–arm interactions, leading to inconsistencies with real-world physical dynamics.

To address these challenges, we propose MTV-World, an embodied world model that introduces Multi-view Trajectory-Video control for accurate and physically grounded visuomotor prediction, as illustrated in Figure~\ref{fig1}.
To tackle the first challenge (unrealistic arm trajectories in predicted videos), we introduce trajectory videos as explicit motion control signals. The generation of trajectory videos consists of three steps:
(1) we map raw action sequences (joint positions) generated by a vision-language-action (VLA) model into Cartesian-space end-effector poses;
(2) we project these end-effector poses into image-space pixel coordinates using calibrated camera intrinsic and extrinsic parameters;
and (3) we render glowing points at the projected coordinates and synthesize temporally continuous trajectory videos that visualize motion evolution over time.

However, projecting 3D actions onto 2D images inevitably leads to a loss of spatial information~\cite{zhen2025tesseract}, which in turn causes inconsistencies with real-world physical interactions (the second major challenge mentioned above).
To alleviate this issue, we introduce two key components:
(1) an object mask as a foreground prior, and
(2) a multi-view framework for spatial consistency.
For the first component, to explicitly model the interaction between the robotic arms and manipulated objects, we utilize an object mask from the initial frame as a foreground prior~\cite{feng2025videoorion}. These masks are automatically obtained through a pipeline that leverages vision language model~(VLM) and referring video object segmentation~(RVOS)~\cite{liang2023local,chen2025wavecl,cuttano2025samwise,yuan2024losh}.
For the second component, the multi-view framework ensures geometric consistency and captures object interactions from complementary perspectives.
With these designs, MTV-World takes the multi-view trajectory videos as input and is conditioned on an initial frame per view to forecast future frames with accurate motion and consistent physical interactions.

Furthermore, to systematically evaluate both robotic motion precision and object interaction accuracy, we develop an evaluation pipeline that reuses the foreground prior extraction strategy described above for perceptual judgment. To quantify spatial consistency, we formulate the evaluation as an object location matching problem and adopt the Jaccard Index between masks from the ground-truth and predicted videos as the evaluation metric~\cite{khoreva2018video,seo2020urvos}. Additionally, since the segmentation performance of RVOS is also influenced by object semantics, it imposes higher demands on the semantic consistency of world models.
Extensive experiments demonstrate that MTV-World achieves state-of-the-art performance in control accuracy and physical interaction consistency.

The major contributions of this work are summarized as follows:
\begin{itemize}
\item We present MTV-World, an embodied world model that introduces Multi-view Trajectory–Video control for accurate and physically consistent visuomotor forecasting.
\item MTV-World utilizes trajectory videos generated from calibrated camera parameters and Cartesian-space mappings as control representations.
\item A multi-view architecture is designed to mitigate spatial information loss and enhance consistency with real-world physical interactions.
\item An evaluation pipeline is established to assess both motion precision and object interaction accuracy via object-location matching and the Jaccard Index.
\end{itemize}

\section{Related Work}
\label{Related Work}

\subsection{World Models}

Recent advances in video generation have significantly enhanced the capacity of models to simulate and understand the physical world~\cite{wan2025wan,zheng2024open,blattmann2023stable,yang2024cogvideox,du2023learning,guo2024prediction}.
World models aim to predict future states from past observations and control actions, providing a foundation for embodied intelligence.
They have been widely applied in simulated environments such as games~\cite{zhang2025matrix,hafner2025mastering} and autonomous driving~\cite{gao2024vista,zhu2025other}.
Several recent works extend video generative models to interactive, controllable settings.
Pandora~\cite{xiang2024pandora} introduces a hybrid autoregressive–diffusion framework capable of simulating world dynamics and executing real-time text-driven control.
Genie~\cite{bruce2024genie} learns a latent action interface from Internet videos, enabling the creation of interactive, playable environments directly from diverse visual-text prompts.
RoboDreamer~\cite{zhou2024robodreamer} factorizes video generation into compositional primitives conditioned on language, allowing flexible task composition for robotic manipulation.
Bar \textit{et al.}~\cite{bar2025navigation} focus on controllable video prediction conditioned on navigation commands, improving spatial reasoning and trajectory planning.
iVideoGPT~\cite{wu2024ivideogpt} generalizes this paradigm into a scalable transformer architecture that integrates visual observations, actions, and rewards into tokenized sequences for next-token prediction.
RoboScape~\cite{shang2025roboscape} incorporates physics-informed supervision via temporal depth prediction and keypoint dynamics, enhancing physical realism in video generation.
The developments of world models highlight the growing potential of generative video models as general-purpose predictive simulators.

\subsection{Embodied World Models}

Beyond open-world video generation, embodied world models aim to predict and interact with the physical world through visual observations and actions in the context of embodied intelligence.
Cosmos~\cite{agarwal2025cosmos} introduces a hierarchical simulation framework that integrates instruction controllability with physical consistency.
Zhu \textit{et al.}~\cite{zhu2025unified} leverage both video and action signals for policy learning, capturing temporal dynamics as a powerful pretraining paradigm for multi-task data.
EnerVerse-AC~\cite{jiang2025enerverse} designs a multi-level action-conditioning mechanism and ray-map encoding for dynamic multi-view image generation, improving generalization through diverse failure trajectories.
FLARE~\cite{zheng2025flare} aligns latent features between diffusion transformers and future observations, enabling policy learning with long-horizon prediction.
Wang \textit{et al.}~\cite{wang2025learning} focus on large-scale video–action modeling for generating high-quality robot learning data.
Li \textit{et al.}~\cite{li2025unified} optimize video and action prediction through a shared latent space that bridges visual and action domains, enhancing inference efficiency and accuracy.
WorldVLA~\cite{cen2025worldvla} unifies VLA modeling and world simulation within a single autoregressive framework to jointly predict actions and future visual outcomes.
Ctrl-World~\cite{guo2025ctrl} extends this direction to a controllable multi-view setting, enabling policy-in-the-loop rollouts purely within imagination space.
TesserAct~\cite{zhen2025tesseract} introduces a 4D embodied video dataset with depth and surface normals for physically consistent manipulation modeling.
RLVR-World~\cite{wu2025rlvr} further combines reinforcement learning with verifiable rewards to directly optimize world models based on quantitative interaction metrics.
World-Eval~\cite{li2025worldeval} provides an evaluation pipeline that encodes low-level actions into latent control representations to assess real-world robot policies online.

\section{Method}
\label{Method}

\subsection{Problem Formulation}

The objective is to construct a world model capable of predicting and interacting with
the physical world through visual observations and actions produced by a generalist robotic policy.
As for a policy $\pi$ (\textit{e.g.}, $\pi_{0}$~\cite{black2024pi_0}), it takes multi-view visual observations and a language instruction as inputs, and outputs a sequence of control actions.
At each time step $t$, the robot observation is represented as $o_t = [I^t_1, \dots, I^t_V, q_t]$, where $[I^t_1, \dots, I^t_V]$ denote the $V$ camera views and $q_t$ corresponds to the robot’s pose (\textit{e.g.}, joint poses).
Conditioned on the current observation and an instruction $l$, the policy produces an action chunk of length $T$:
\begin{equation}
a_{t+1}, a_{t+2}, \dots, a_{t+T} \sim \pi(\cdot \mid o_t, l).
\end{equation}
The world model $W$ aims to simulate the interaction of the physical environment by predicting the perceptual outcomes resulting from executing this action sequence $A_t = [a_{t+1}, \dots, a_{t+T}]$, along with the language instruction $l$.
Formally, $W$ generates the corresponding future observations for each step:
\begin{equation}
o_{t+1}, \dots, o_{t+T} \sim W(\cdot \mid o_t, A_t, l).
\end{equation}

\subsection{Trajectory Representation}
\begin{figure*}[htbp]
\centering
\includegraphics[width=1\linewidth]{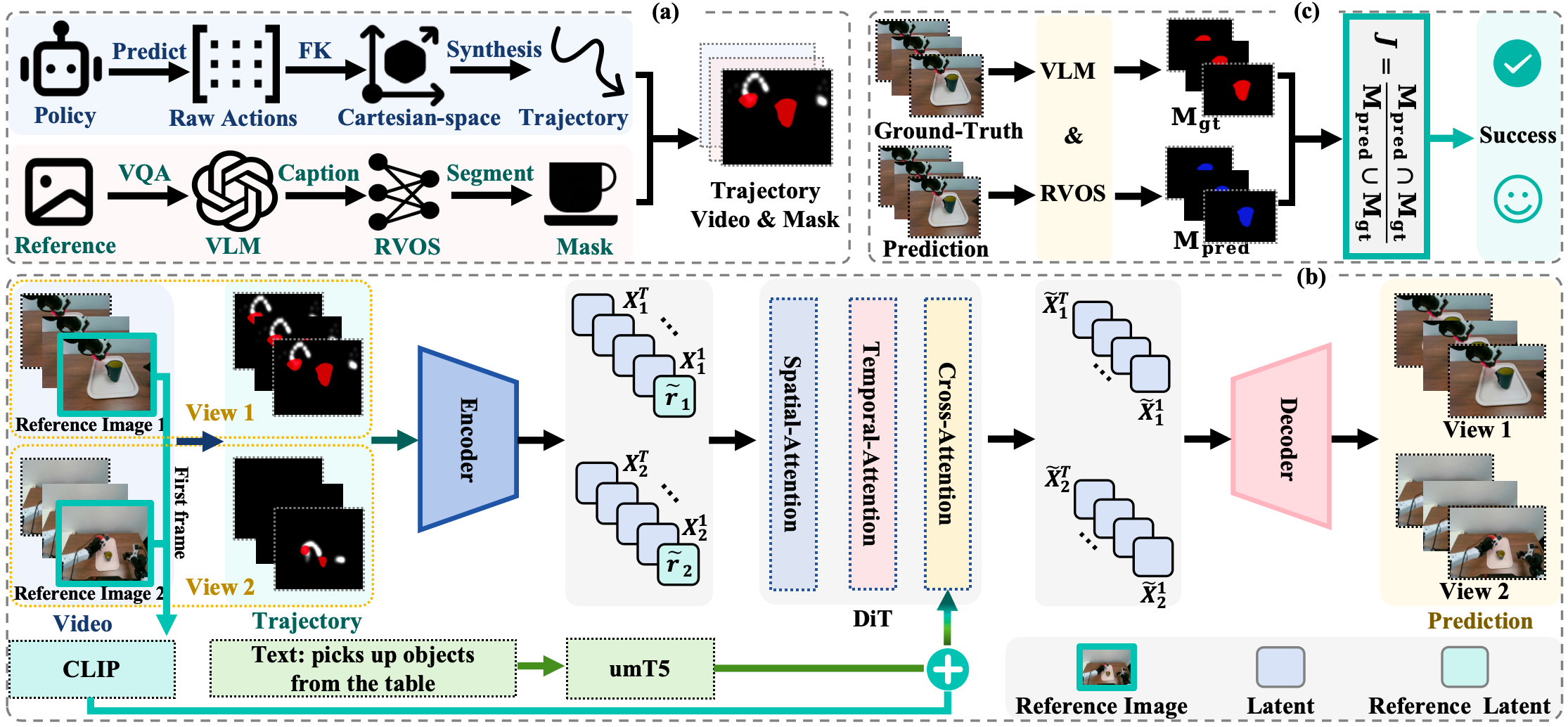}
\caption{
Illustration of the MTV-World framework. (a) Trajectory and object representation: trajectory control videos are generated by combining object masks and the trajectories of luminous points on the image to serve as control inputs. (b) MTV-World architecture: the model takes multi-view video sequences and trajectory control videos as inputs. The first frame is used as a reference image, which is simultaneously processed by CLIP for semantic encoding and by a shared VAE encoder to obtain reference latents, which are later removed before decoding. (c) Evaluation pipeline: performance is assessed by measuring the spatial alignment between the predicted and ground-truth object masks in videos using the Jaccard Index.}
\label{fig2}
\end{figure*}

To tackle the challenge of unrealistic arm trajectories in predicted videos, we introduce trajectory videos as explicit motion control signals, as illustrated in Figure~\ref{fig2}~(a).
We first calibrate each camera to obtain its intrinsic parameters, including focal lengths $f_x$, $f_y$ and principal point $c_x$, $c_y$, as well as its extrinsic parameters, represented by the camera-to-base rotation matrix $R$ and translation vector $t$. We then transform the robot arm joint positions $q_t$ into their corresponding end-effector poses in Cartesian space. Specifically, we use the forward kinematics (FK) model of a 6 degrees of freedom (6-DoF) robot defined by its Denavit–Hartenberg parameters to compute the 3D position of each point, denoted as \(p_w = (x_w, y_w, z_w)\).

We project this 3D point from the robot’s base (world) coordinate system onto the 2D pixel location on the image plane. To do so, \(p_w\) is first transformed into the camera coordinate system \(p_c\) using the rotation matrix \(R\) and translation vector \(t\) obtained from camera calibration:
\begin{equation}
p_c = R^\top (p_w - t).
\end{equation}
The resulting camera coordinates $p_c = (x_c, y_c, z_c)$ are then projected onto the image plane through the pinhole camera model:
\begin{equation}
u = f_x \frac{x_c}{z_c} + c_x, \quad v = f_y \frac{y_c}{z_c} + c_y,
\end{equation}
where \( f_x, f_y \) denote the focal lengths, \( (c_x, c_y) \) the principal point offsets encoded in the intrinsic matrix \( K \), and \( (u, v) \) represent the projected pixel coordinates.

Finally, we synthesize trajectory videos for dual-arm motions. For each frame, the 3D end-effector positions are projected onto 2D pixel coordinates \((u, v)\). These projected points are drawn as fading trails of glowing points to demonstrate the continuity of motion. The frames are then composed into a single-view trajectory video \(X^\text{traj}_v\), providing an intuitive visualization of the control policy. For the multi-view setting, this process is performed for each camera view, producing \(V\) synchronized trajectory videos \(X^\text{traj}\).

\subsection{Object Representation}
\label{method:obj-represent}
To explicitly model the interaction between the robot arm and manipulated objects, we introduce an object mask from the initial frame as a foreground prior. To automatically obtain these masks, we adopt a pipeline inspired by RVOS. Specifically, we first generate textual descriptions for each object by prompting a VLM in a visual question answering (VQA) manner. Given an input image, the VLM produces a set of object descriptions, which are then provided to the RVOS model to generate segmentation masks for all objects across the entire video. The resulting full-video masks are later used for automatic evaluation, as detailed in Section~\ref{method:eval-pipe}.

Since the world model is defined to predict future frames conditioned on the first observed frame, we utilize only the object mask extracted from the initial frame as the foreground prior. Finally, object masks in a single frame are duplicated across all frames to construct the reference mask sequence for the entire video.

\subsection{Model Architecture}

The overall model architecture is illustrated in Figure~\ref{fig2}~(b).
We adopt the first frame as the reference image to guide the video synthesis process. Specifically, given $V$ synchronized camera views, the video set is denoted as \(X^\text{video} \in \mathbb{R}^{V \times (1+T) \times 3 \times h \times w}\), where $V$ indicates the number of views, $1$ denotes the reference frame, and $T$ denotes the number of following frames; $h$ and $w$ represent the frame height and width, respectively. MTV-World aims to predict future $T$ frames for all views, conditioned on the observed initial frames and their corresponding control sequences. Both the multi-view video sequences $X^\text{video}$ and the trajectory control videos $X^\text{traj}$ are encoded by a shared variational autoencoder~(VAE)~\cite{wu2025improved} encoder into compact latent representations. Finally, the reference latent and the video latent are concatenated along the channel dimension to form the unified video latent representation $X$ used for the subsequent diffusion process.

To provide scene priors across multiple views, MTV-World incorporates $V$ reference images $\{I_v\}_{v=1}^V$ and a text prompt $l$. Each reference image is encoded by a pretrained CLIP~\cite{radford2021learning} image encoder $\text{CLIP}(\cdot)$ to extract global semantic features:
\(
F_v = \text{CLIP}(I_v).
\)
These features are projected into the transformer’s embedding dimension via a learned linear projection. The text prompt $l$ is tokenized and embedded by a umT5~\cite{chung2023unimax} text encoder $\text{umT5}(\cdot)$, yielding contextual tokens:
\(
F_\text{text} = \text{umT5}(l).
\)
Finally, the projected CLIP embeddings from all views are concatenated together with the text embeddings to form a unified multimodal context:
\(
\{F_1, \dots, F_V, F_\text{text}\}.
\)
This unified multimodal context subsequently interacts with the latents through cross-attention within the diffusion transformer (DiT)~\cite{peebles2023scalable} to ensure the incorporation of scene priors.

To ensure multi-view appearance consistency throughout the generation process, each reference image is also processed by the same VAE encoder used in the video encoding stage: \(r_v = \text{Enc}(I_v).\)
Here, $\text{Enc}(\cdot)$ denotes the VAE encoder. The encoded latent feature $r_v$ is passed through a lightweight adapter $\phi(\cdot)$ to align its channel dimension with the latents as: \(\tilde{r}_v = \phi(r_v), \quad \tilde{r}_v \in \mathbb{R}^{c'\times 1 \times h' \times w'}, \)
where \(c'\), \(h'\), and \(w'\) represent the channel, height, and width dimensions of the latent feature, respectively. 
The processed reference latents from all $V$ views, denoted as $\{\tilde{r}_v\}_{v=1}^V$, are integrated into the latent sequence to provide cross-view appearance constraints. Specifically, for each view $v$, the reference latent $\tilde{r}_v$ is concatenated with its corresponding latent sequence $X_v = [X_v^1, X_v^2, \dots, X_v^{T}]$ along the temporal dimension. The resulting multi-view latent set can be represented as:
\begin{equation}
X = 
\begin{bmatrix}
\tilde{r}_1 & X_1^1 & X_1^2 & \dots & X_1^{T} \\
\vdots & \vdots & \vdots & \ddots & \vdots \\
\tilde{r}_V & X_V^1 & X_V^2 & \dots & X_V^{T}
\end{bmatrix},
\end{equation}
where each row corresponds to one camera view.

After processing through all diffusion transformer layers, the output sequence $\hat{X} = \text{DiT}(X)$ preserves both the reference and generated tokens. To reconstruct the latent sequences, we remove the reference tokens and retain only the frame-wise outputs:
\(
X_{\text{kept}} = [\hat{X}_1^{1}, \dots, \hat{X}_1^{T}; \dots; \hat{X}_V^{1}, \dots, \hat{X}_V^{T}].
\)
The retained tokens correspond to the predicted latent representations:
\(
X_{\text{kept}} \in \mathbb{R}^{V \times c' \times T \times h' \times w'},
\)
which are decoded into the pixel space via the shared VAE decoder $\text{Dec}(\cdot)$:
\(
X^\text{pred} = \text{Dec}(X_{\text{kept}}),
\)
yielding multi-view video predictions $\{X^\text{pred}_v\}_{v=1}^V$ that maintain semantic alignment and visual consistency across all views.

\subsection{Evaluation Pipeline}
\label{method:eval-pipe}
Existing evaluation methods for world models primarily rely on image- or video-based metrics such as Fréchet Inception Distance (FID)~\cite{heusel2017gans} and Fréchet Video Distance (FVD)~\cite{unterthiner2018towards}, which mainly measure perceptual quality rather than control accuracy. However, these metrics fail to measure the accuracy of object–arm interactions or object movement.

To address these issues, we design a new automated evaluation pipeline that directly measures interactions and object movement accuracy. Specifically, we hypothesize that if a generated video accurately reproduces the robot’s manipulation and the corresponding object interactions, then the predicted object positions should closely match those in the ground-truth video.

We formulate this positional consistency as a mask-matching problem using RVOS. The segmentation masks of objects are automatically obtained for both generated and real videos. For each frame \(t\) in a video, we compute the frame-level Jaccard Index, and the overall video-level Jaccard Index is obtained by averaging across all frames:
\begin{equation}
\mathcal{J}_t = \frac{M_\text{pred}^t \cap M_\text{gt}^t}{M_\text{pred}^t \cup M_\text{gt}^t}, \quad
\mathcal{J} = \frac{1}{T} \sum_{t=1}^{T} \mathcal{J}_t,
\end{equation}
where \( M_\text{pred}^t \) and \( M_\text{gt}^t \) denote the predicted and ground-truth masks at frame \(t\), respectively, and \(T\) is the total number of frames in the video. 

The complete evaluation pipeline is fully automated, with the overall workflow depicted in Figure~\ref{fig2}~(c). Following the procedure described in Section~\ref{method:obj-represent}, we employ a VLM in a VQA manner to automatically generate text descriptions for each object. These descriptions are then fed into an RVOS model to produce object segmentation masks across the entire video. The resulting masks are used to compute the $\mathcal{J}$ scores between generated and ground-truth videos, thereby enabling an end-to-end quantitative evaluation of world model performance on object-centric consistency and manipulation accuracy.

\section{Experiments}
\label{Expe}

\subsection{Dataset and Task Setup}

\noindent
\textbf{Dataset Overview.}
The experiments are conducted on a self-collected dual-arm robotic manipulation dataset containing a total of 1,492 videos across 15 real-world tasks. Each video records a complete execution sequence of two robotic arms performing coordinated manipulation behaviors. The dataset includes both successful and failed trials, with an approximate success-to-failure ratio of 8:2. In failure cases, the gripper may miss the target object but still execute the full trajectory. Introducing these failure cases further tests the model’s ability to accurately determine whether the interaction was successful.

The dataset is split into training and testing sets with a ratio of 5:5, corresponding to 748 training samples and 744 testing samples. Additionally, we collected 395 extra videos involving only random pick-and-place tasks, which are not part of the aforementioned 15 tasks, to test the model's zero-shot capability.

\smallskip \noindent
\textbf{Task Setup.}
The dataset comprises 15 distinct manipulation tasks adapted from the
RoboTwin benchmark~\cite{mu2025robotwin,chen2025robotwin}: (1) Bussing Table, (2) Collect Food, (3) Collect Tableware, (4) Collect Toy, (5) Move and Stack Block AB, (6) Move and Stack Plate, (7) Move Object Two, (8) Place Block A2B Left, (9) Place Block A2B Right, (10) Place Block AB2C Left, (11) Place Block AB2C Right, (12) Place Bread Plate, (13) Place Cup Plate, (14) Shake Bottle, and (15) Stack Blocks Two. 

Each task contains 97–101 videos to ensure balanced coverage. Each video includes at least two manipulable objects, excluding trays. The dataset encompasses a diverse range of object categories, such as colorful toy bricks of various sizes (blocks); plates, bowls, and cups (tableware); food items like bread, carrots, eggplants, and corn; as well as trays, which serve as collection targets or support surfaces. These tasks involve a wide range of motion primitives, including grasping, stacking, shaking, collecting, and spatial rearrangement, thereby providing a diverse benchmark for evaluating embodied manipulation models.

\smallskip \noindent
\textbf{Robot Setup.}
We employ the ALOHA-style dual-arm robotic platform, named YAM, which consists of two manipulators, each with 6-DoF. This configuration results in a combined 14-dimensional state and action space for bimanual control. Two RealSense D457 cameras are mounted at different positions: one at a relatively lower position (View 1) and another at a relatively higher position (View 2) to capture both close-up and global views of the manipulation scene, as illustrated in Figure~\ref{fig1}.

\smallskip \noindent
\textbf{Implementation Details.}
The proposed world model is implemented based on the Wan 2.1 framework~\cite{wan2025wan}. The model is trained with an initial learning rate of 1$\times10^{-4}$ and a batch size of 8 using eight NVIDIA H20 GPUs. We adopt LoRA~\cite{hu2022lora} fine-tuning with a rank of 32. The training process runs for 8 epochs. Each training sample consists of two camera views, each containing 81 frames, resulting in a total of 81 $\times$ 2 frames per sample at a resolution of 288 $\times$ 384. The text prompt is set as ``The robotic arm picks up objects from the table and places them'' for all tasks. We adopt Doubao Thinking Vision~\cite{guo2025seed1} as the vision–language model, and SAMWISE~\cite{cuttano2025samwise} as the referring video object segmentation model.

\smallskip \noindent
\textbf{Evaluation Criteria.}
We adopt FID~\cite{heusel2017gans} and FVD~\cite{unterthiner2018towards} to assess the perceptual quality of the generated videos. We also employ the evaluation pipeline introduced in Section~\ref{method:eval-pipe}, which measures temporal and spatial consistency using the Jaccard Index $\mathcal{J}$~\cite{khoreva2018video,seo2020urvos}.

\subsection{Baseline Comparison}

\begin{table}[ht]
\centering
\small
\resizebox{0.48\textwidth}{!}{%
\setlength{\tabcolsep}{3pt}
\begin{tabular}{c|ccc|ccc}
\toprule[1pt]
\multirow{2}{*}{Method} & \multicolumn{3}{c|}{View 1} & \multicolumn{3}{c}{View 2} \\
 & FID $\downarrow$ & FVD $\downarrow$ & $\mathcal{J}$ $\uparrow$ & FID $\downarrow$ & FVD $\downarrow$ & $\mathcal{J}$ $\uparrow$ \\
\midrule
World-Eval~\cite{li2025worldeval} & -- & -- & -- & 39.0 & 518.2 & 18.3 \\
Ctrl-World~\cite{guo2025ctrl} & 31.7 & 162.5 & 35.3 & 33.7 & 263.1 & 22.2 \\
Fast-Token~\cite{pertsch2025fast} & 33.5 & 198.0 & 36.7 & 41.5 & 216.8 & 26.3 \\
MTV-World (Ours) & \textbf{22.0} & \textbf{42.8} & \textbf{54.9} & \textbf{17.5} & \textbf{65.8} & \textbf{45.0} \\
\bottomrule[1pt]
\end{tabular}
}
\caption{Comparison with baseline methods on the Self-collected Dataset.}
\label{tab:self_collected}
\end{table}

We compare MTV-World with three representative baselines characterized by their diverse action representation strategies: 
(1) Vector-based approach, World-Eval~\cite{li2025worldeval}, which utilizes a Policy2Vec strategy to encode raw action sequences into latent action vectors; 
(2) 3D-projection-based approach, Ctrl-World~\cite{guo2025ctrl}, which employs an MLP adapter for raw action (3D end-effector positions) processing; and 
(3) Token-based approach, Fast-Token~\cite{pertsch2025fast}, which uses the Discrete Cosine Transform to compress actions into tokens. 
Note that because World-Eval only supports single-view training and inference, its evaluation is restricted to View 2. The evaluation metrics include Fréchet Inception Distance (FID) for visual quality, Fréchet Video Distance (FVD) for temporal consistency, and the Jaccard Index ($\mathcal{J}$) for spatial interaction consistency.

\smallskip \noindent
\textbf{Self-collected Dataset.}
As shown in Table~\ref{tab:self_collected}, the proposed MTV-World significantly outperforms all baselines across all metrics. By converting raw actions into multi-view trajectory control videos, our approach provides a more explicit and physically grounded representation of robot behaviors compared to standard projections or latent action encodings.

\begin{table*}[ht]
\centering
\small
\resizebox{0.65\textwidth}{!}{%
\setlength{\tabcolsep}{4pt}
\begin{tabular}{c|ccc|ccc|ccc}
\toprule[1pt]
\multirow{2}{*}{Method} & \multicolumn{3}{c|}{Magazines} & \multicolumn{3}{c|}{Pen Holder} & \multicolumn{3}{c}{TV Cabinet} \\
 & FID $\downarrow$ & FVD $\downarrow$ & $\mathcal{J}$ $\uparrow$ & FID $\downarrow$ & FVD $\downarrow$ & $\mathcal{J}$ $\uparrow$ & FID $\downarrow$ & FVD $\downarrow$ & $\mathcal{J}$ $\uparrow$ \\
\midrule
Ctrl-World~\cite{guo2025ctrl} & 30.0 & 438.3 & 26.4 & 31.6 & 242.1 & 29.1 & 44.6 & 439.0 & 27.4 \\
Fast-Token~\cite{pertsch2025fast} & 30.8 & 387.2 & 28.1 & 38.0 & 297.9 & 28.1 & 39.2 & 390.1 & 33.8 \\
MTV-World (Ours) & \textbf{26.2} & \textbf{206.6} & \textbf{41.8} & \textbf{22.5} & \textbf{130.1} & \textbf{38.2} & \textbf{36.0} & \textbf{195.3} & \textbf{43.2} \\
\bottomrule[1pt]
\end{tabular}
}
\caption{Comparison with baseline methods on the RealSource-World Dataset.}
\label{tab:realsource}
\end{table*}

\smallskip \noindent
\textbf{RealSource-World Dataset.}
To validate the generalization capability of our approach, we further evaluated MTV-World on RealSource-World\footnote{\scriptsize \url{https://huggingface.co/datasets/RealSourceData/RealSource-World}}, a large-scale dataset that provides the requisite camera intrinsic and extrinsic parameters. We selected three distinct manipulation tasks (Magazines, Pen Holder, and TV Cabinet) containing a total of 600 videos, split into 75\% for training and 25\% for testing. Since RealSource-World is a single-view dataset, we selected the high camera view for both training and testing phases.

As reported in Table~\ref{tab:realsource}, MTV-World significantly outperforms the Ctrl-World and Fast-Token baselines on this benchmark. The substantial improvements confirm that our trajectory-video control mechanism maintains high consistency and accurate object-arm interactions.

\begin{table*}[ht]
\centering
\small
\setlength{\tabcolsep}{4pt}
\resizebox{\textwidth}{!}{%
\begin{minipage}{1.1\textwidth}

\begin{tabular*}{\linewidth}{@{} l @{\extracolsep{\fill}} cc|cc|cc|cc|cc|cc @{}}
\toprule
\multirow{2}{*}{Method} &
\multicolumn{2}{c|}{Bussing Table} &
\multicolumn{2}{c|}{Collect Food} &
\multicolumn{2}{c|}{Collect Tableware} &
\multicolumn{2}{c|}{Collect Toy} &
\multicolumn{2}{c|}{Move Block AB} &
\multicolumn{2}{c}{Move Plate} \\
& Low & High & Low & High & Low & High & Low & High & Low & High & Low & High \\
\midrule
World-Eval~\cite{li2025worldeval} & -- & 19.9 & -- & 15.3 & -- & 22.8 & -- & 17.6 & -- & 10.8 & -- & 13.8 \\
Ctrl-World~\cite{guo2025ctrl} & 36.9 & 34.4 & 25.9 & 19.2 & 29.5 & 19.2 & 34.9 & 26.8 & 22.5 & 13.5 & 27.6 & 20.2 \\
Fast-Token~\cite{pertsch2025fast} & 34.2 & 28.6 & 31.7 & 22.9 & 30.5 & 20.6 & 35.7 & 26.1 & 25.6 & 17.2 & 25.9 & 19.6 \\
Zero-Shot & 50.7 & 50.0 & 47.4 & 39.9 & 44.0 & 37.4 & 49.5 & 42.4 & 30.6 & 20.7 & 36.0 & 31.1 \\
Single-View & 51.1 & 43.8 & 50.0 & 38.8 & 43.2 & 36.7 & 53.8 & 43.2 & 33.2 & 21.2 & 36.9 & 31.1 \\
w/o Mask & 57.7 & 56.8 & 49.6 & 43.8 & 51.4 & 40.5 & 56.0 & 48.6 & 40.1 & 26.0 & 54.2 & 41.7 \\
MTV-World & 55.7 & 54.7 & 50.7 & 47.6 & 52.4 & 44.8 & 56.1 & 50.5 & 38.7 & 27.2 & 48.9 & 39.6 \\
\bottomrule
\end{tabular*}

\vspace{0.6em}

\begin{tabular*}{\linewidth}{@{} l @{\extracolsep{\fill}} cc|cc|cc|cc|cc @{}}
\toprule
\multirow{2}{*}{Method} &
\multicolumn{2}{c|}{Move Object Two} &
\multicolumn{2}{c|}{Block A2B Left} &
\multicolumn{2}{c|}{Block A2B Right} &
\multicolumn{2}{c|}{Block AB2C Left} &
\multicolumn{2}{c}{Block AB2C Right} \\
& Low & High & Low & High & Low & High & Low & High & Low & High \\
\midrule
World-Eval~\cite{li2025worldeval} & -- & 12.0 & -- & 16.6 & -- & 19.9 & -- & 19.7 & -- & 17.2 \\
Ctrl-World~\cite{guo2025ctrl} & 25.4 & 18.6 & 37.4 & 13.4 & 38.4 & 15.5 & 36.9 & 23.3 & 31.7 & 19.0 \\
Fast-Token~\cite{pertsch2025fast} & 26.9 & 21.8 & 36.1 & 26.2 & 44.4 & 29.5 & 32.9 & 22.8 & 39.3 & 31.1 \\
Zero-Shot & 33.5 & 28.3 & 57.6 & 38.9 & 55.5 & 39.1 & 43.4 & 36.2 & 51.0 & 43.2 \\
Single-View & 31.9 & 27.6 & 64.3 & 42.8 & 60.5 & 39.5 & 47.3 & 37.7 & 52.3 & 40.3 \\
w/o Mask & 37.2 & 30.8 & 58.0 & 40.4 & 59.9 & 42.8 & 49.0 & 39.2 & 53.9 & 43.6 \\
MTV-World & 38.6 & 32.6 & 63.5 & 46.5 & 59.8 & 46.0 & 53.1 & 43.0 & 53.8 & 45.4 \\
\bottomrule
\end{tabular*}
\vspace{0.6em}

\begin{tabular*}{\linewidth}{@{} l @{\extracolsep{\fill}} cc|cc|cc|cc|cc @{}}
\toprule
\multirow{2}{*}{Method} &
\multicolumn{2}{c|}{Place Bread Plate} &
\multicolumn{2}{c|}{Place Cup Plate} &
\multicolumn{2}{c|}{Shake Bottle} &
\multicolumn{2}{c|}{Stack Blocks Two} &
\multicolumn{2}{c}{Task Averaged} \\
& Low & High & Low & High & Low & High & Low & High & Low & High \\
\midrule
World-Eval~\cite{li2025worldeval} & -- & 31.5 & -- & 27.3 & -- & 13.2 & -- & 16.7 & -- & 18.3 \\
Ctrl-World~\cite{guo2025ctrl} & 58.0 & 42.2 & 47.6 & 32.5 & 39.4 & 16.6 & 36.7 & 17.8 & 35.3 & 22.2 \\
Fast-Token~\cite{pertsch2025fast} & 57.1 & 43.4 & 51.9 & 31.9 & 37.8 & 23.6 & 38.8 & 28.0 & 36.7 & 26.3 \\
Zero-Shot & 66.3 & 54.9 & 63.4 & 43.0 & 62.0 & 47.8 & 49.2 & 32.5 & 49.4 & 39.0 \\
Single-View & 65.3 & 47.6 & 63.1 & 41.3 & 59.6 & 40.1 & 55.8 & 38.8 & 51.3 & 38.0 \\
w/o Mask & 69.1 & 57.3 & 64.9 & 43.2 & 62.9 & 43.4 & 57.3 & 41.5 & 54.8 & 42.6 \\
MTV-World & 68.3 & 58.9 & 64.2 & 48.2 & 64.5 & 50.0 & 54.6 & 40.6 & \textbf{54.9} & \textbf{45.0} \\
\bottomrule
\end{tabular*}

\end{minipage}%
} 
\caption{Per-task $\mathcal{J}$ score comparison across two camera views (Low: View 1, High: View 2).}
\label{tab:ablation}
\end{table*}

\subsection{Ablation Study}
We conduct comprehensive ablation experiments to systematically validate the effectiveness of the proposed MTV-World. Specifically, we investigate: (1) the impact of different trajectory representations, (2) the benefits of multi-view input compared to single-view settings, (3) the zero-shot generalization ability across novel views, tasks, and objects, (4) the influence of object masking.

\smallskip \noindent
\textbf{Trajectory Representation.}
To assess the effectiveness of our trajectory-based control, we compare MTV-World against the previously introduced baselines, which represent diverse action conditioning strategies: the vector-based World-Eval~\cite{li2025worldeval}, the 3D-projection-based Ctrl-World~\cite{guo2025ctrl}, and the token-based Fast-Token~\cite{pertsch2025fast}. As shown in Table~\ref{tab:ablation}, MTV-World consistently outperforms these methods across all metrics. This demonstrates that explicit trajectory video control provides superior guidance for motion and interaction generation compared to implicit or compressed action representations.

\smallskip \noindent
\textbf{Multi-View \textit{vs.} Single-View.}
To validate the benefits of multi-view learning, we design a Single-View variant ($V=1$) that utilizes only a single viewpoint during both training and inference. The comparison results are presented in Table~\ref{tab:ablation}. In the relatively simple low-view scenario, the Single-View model achieves a slightly lower $\mathcal{J}$ of 51.3. However, in the more complex high-view scenario, its performance drops significantly to 38.0, largely lagging behind the Multi-View model (45.0). Notably, the Single-View approach requires training and deploying two separate models (one for View 1 and another for View 2), whereas MTV-World achieves superior multi-view consistency and performance through a single, unified framework.

\smallskip \noindent
\textbf{Zero-Shot Ability.}
MTV-World also exhibits strong zero-shot generalization. To evaluate this capability, we introduce an additional zero-shot dataset containing 395 samples, covering three challenging settings: (1) View, where entirely novel camera positions and orientations are used; (2) Task, where unseen manipulation tasks (\textit{e.g.}, random pick-and-place) are introduced; and (3) Object, where 50\% objects do not appear in the training or testing data. As shown in Table~\ref{tab:ablation}, the zero-shot model achieves competitive results with 49.4 $\mathcal{J}$ for View~1 and 39.0 $\mathcal{J}$ for View~2. The results demonstrate remarkable generalization to unseen viewpoints, tasks, and objects.

\smallskip \noindent
\textbf{Object Mask.}
We examine the effect of the object mask on modeling the interaction between the robot arm and the manipulated objects. As shown in Table~\ref{tab:ablation}, the model with the object mask achieves higher $\mathcal{J}$ scores across both views, demonstrating its effectiveness in capturing contact-aware motion dynamics.

\begin{figure*}[htbp]
\centering
\includegraphics[width=0.9\linewidth]{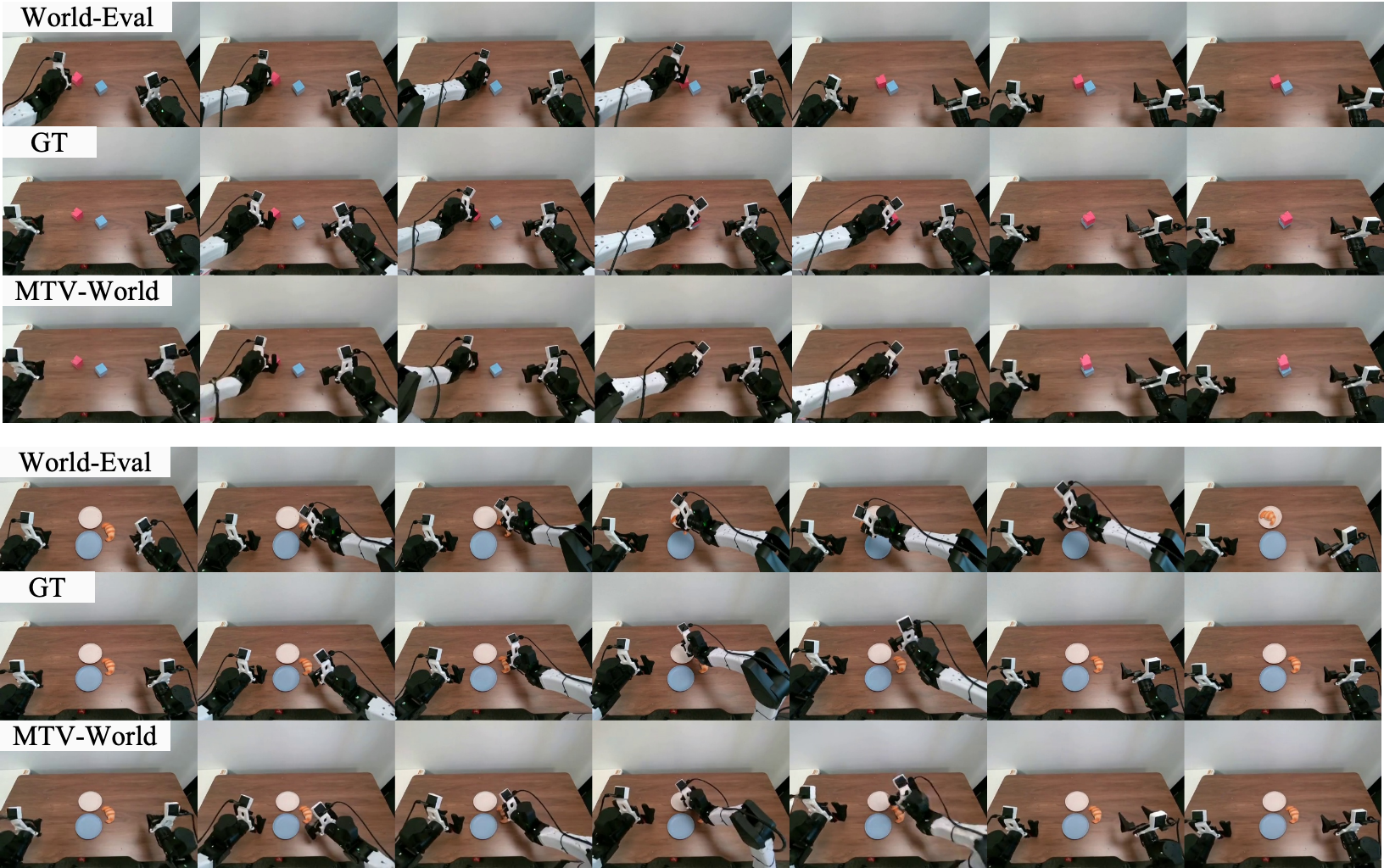}
\caption{
Qualitative comparison with baseline methods on the self-collected dataset. Two representative examples are shown: a successful Stack Blocks task (top) and a failed rollout of the Place Bread Plate task (bottom).}
\label{fig3}
\end{figure*}

\begin{figure*}[htbp]
\centering
\includegraphics[width=0.9\linewidth]{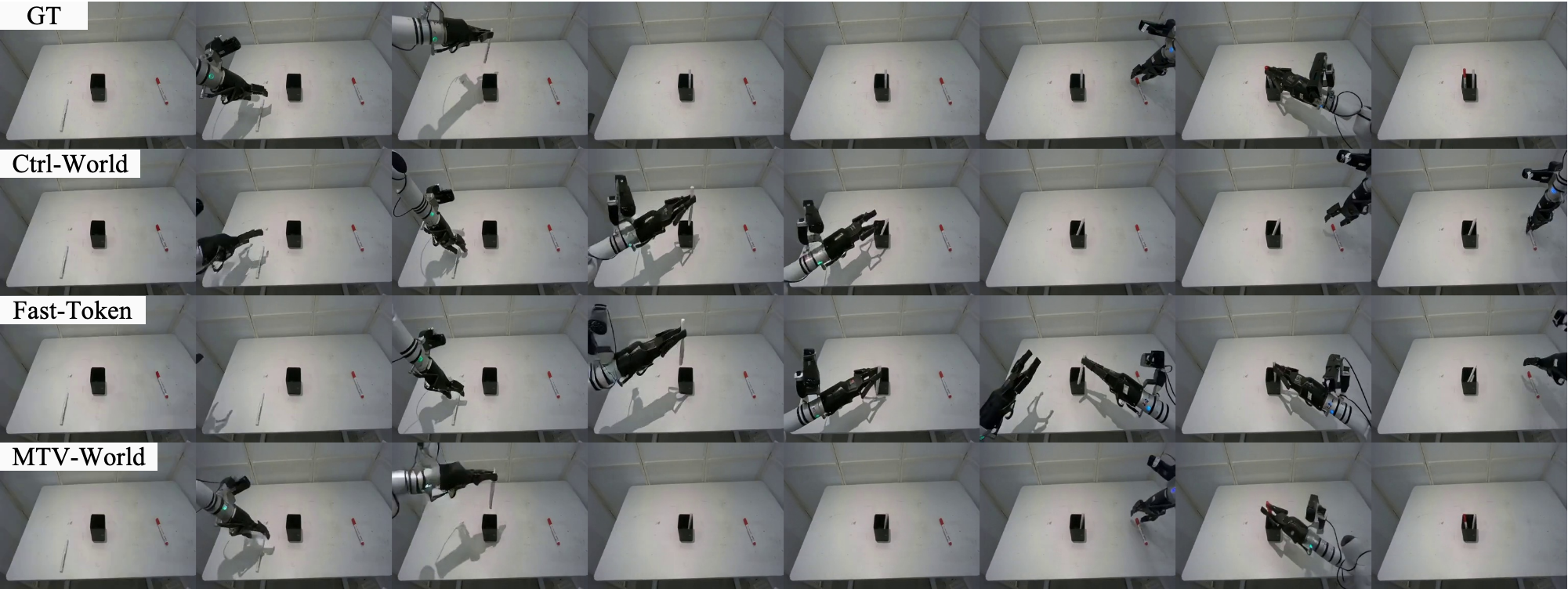}
\caption{
Qualitative comparison with baseline methods on the RealSource-World (Pen Holder task).}
\label{fig4}
\end{figure*}

\begin{figure*}[htbp]
\centering
\includegraphics[width=0.9\linewidth]{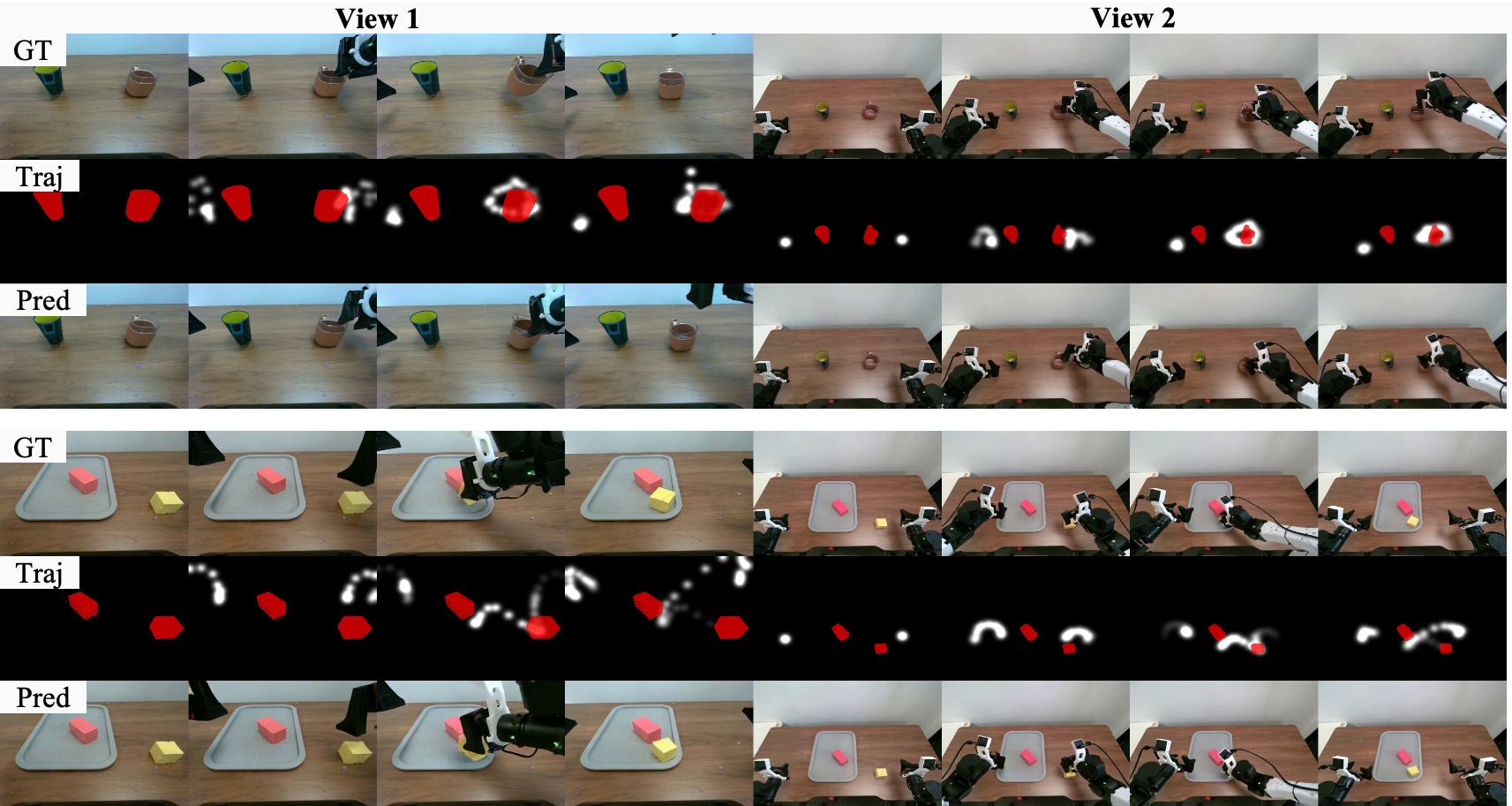}
\caption{
Visualization of the generated multi-view trajectory videos for two representative examples: Shake Bottle (top) and Collect Toy (bottom).}
\label{fig5}
\end{figure*}

\subsection{Visualization Results}

\smallskip \noindent
\textbf{Self-collected Dataset.}
We present qualitative comparisons with the baseline model World-Eval on the self-collected dataset in Figure~\ref{fig3}, highlighting two representative examples of successful and failed rollouts. The first example corresponds to a successful Stack Blocks task, where World-Eval mistakenly predicts placing the block beside the target. In contrast, MTV-World correctly predicts the stacking action, consistent with the ground truth. The second example illustrates a failed rollout in the Place Bread Plate task, where the ground-truth video demonstrates that the bread was not successfully placed onto the plate. In this scenario, World-Eval incorrectly generates a successful placement, whereas MTV-World faithfully reproduces the failed trajectory, accurately reflecting the true physical outcome.

\smallskip \noindent
\textbf{RealSource-World Dataset.}
We also present qualitative comparisons with Ctrl-World and Fast-Token on the RealSource-World dataset for the Pen Holder task, as shown in Figure~\ref{fig4}. As observed, the predictions generated by our method align closely with the ground truth. Notably, MTV-World is the only model among the three evaluated methods capable of successfully completing the task, without inconsistencies in robotic arm movements or interaction failures.

\smallskip \noindent
\textbf{Multi-View Trajectory Videos.}
We further visualize the generated multi-view trajectory videos, as illustrated in Figure~\ref{fig5}, for two representative tasks: Shake Bottle (top) and Collect Toy (bottom). The former requires the robot to grasp and shake the cup in a circular motion. From the rendered trajectory videos, we observe that in both camera views, the glowing trajectory points form a circular trail around the object’s initial mask region, indicating consistent spatio-temporal motion prediction. In the Collect Toy example, the trajectory points clearly reveals the robot end-effector approaching and placing the toy, providing an intuitive visualization of the manipulation dynamics.

\section{Conclusion}
In this work, we introduced MTV-World, an embodied world model that leverages Multi-view Trajectory–Video control for precise and physically consistent visuomotor prediction. By representing control signals as trajectory videos and incorporating a multi-view framework with object-aware priors, MTV-World effectively bridges the gap between low-level actions and realistic physical interactions. Furthermore, the evaluation pipeline provides a systematic and objective means to assess both motion precision and interaction accuracy. Extensive experiments demonstrate that MTV-World achieves state-of-the-art performance in complex dual-arm manipulation scenarios.


\begin{thebibliography}{51}
\providecommand{\natexlab}[1]{#1}
\providecommand{\url}[1]{\texttt{#1}}
\expandafter\ifx\csname urlstyle\endcsname\relax
  \providecommand{\doi}[1]{doi: #1}\else
  \providecommand{\doi}{doi: \begingroup \urlstyle{rm}\Url}\fi

\bibitem[Agarwal et~al.(2025)Agarwal, Ali, Bala, Balaji, Barker, Cai, Chattopadhyay, Chen, Cui, Ding, et~al.]{agarwal2025cosmos}
Niket Agarwal, Arslan Ali, Maciej Bala, Yogesh Balaji, Erik Barker, Tiffany Cai, Prithvijit Chattopadhyay, Yongxin Chen, Yin Cui, Yifan Ding, et~al.
\newblock Cosmos world foundation model platform for physical ai.
\newblock \emph{arXiv preprint arXiv:2501.03575}, 2025.

\bibitem[Bar et~al.(2025)Bar, Zhou, Tran, Darrell, and LeCun]{bar2025navigation}
Amir Bar, Gaoyue Zhou, Danny Tran, Trevor Darrell, and Yann LeCun.
\newblock Navigation world models.
\newblock In \emph{Proceedings of the Computer Vision and Pattern Recognition Conference}, pages 15791--15801, 2025.

\bibitem[Black et~al.(2024)Black, Brown, Driess, Esmail, Equi, Finn, Fusai, Groom, Hausman, Ichter, et~al.]{black2024pi_0}
Kevin Black, Noah Brown, Danny Driess, Adnan Esmail, Michael Equi, Chelsea Finn, Niccolo Fusai, Lachy Groom, Karol Hausman, Brian Ichter, et~al.
\newblock $pi\_0$: A vision-language-action flow model for general robot control.
\newblock \emph{arXiv preprint arXiv:2410.24164}, 2024.

\bibitem[Blattmann et~al.(2023)Blattmann, Dockhorn, Kulal, Mendelevitch, Kilian, Lorenz, Levi, English, Voleti, Letts, et~al.]{blattmann2023stable}
Andreas Blattmann, Tim Dockhorn, Sumith Kulal, Daniel Mendelevitch, Maciej Kilian, Dominik Lorenz, Yam Levi, Zion English, Vikram Voleti, Adam Letts, et~al.
\newblock Stable video diffusion: Scaling latent video diffusion models to large datasets.
\newblock \emph{arXiv preprint arXiv:2311.15127}, 2023.

\bibitem[Bruce et~al.(2024)Bruce, Dennis, Edwards, Parker-Holder, Shi, Hughes, Lai, Mavalankar, Steigerwald, Apps, et~al.]{bruce2024genie}
Jake Bruce, Michael~D Dennis, Ashley Edwards, Jack Parker-Holder, Yuge Shi, Edward Hughes, Matthew Lai, Aditi Mavalankar, Richie Steigerwald, Chris Apps, et~al.
\newblock Genie: Generative interactive environments.
\newblock In \emph{Forty-first International Conference on Machine Learning}, 2024.

\bibitem[Cen et~al.(2025)Cen, Yu, Yuan, Jiang, Huang, Guo, Li, Song, Luo, Wang, et~al.]{cen2025worldvla}
Jun Cen, Chaohui Yu, Hangjie Yuan, Yuming Jiang, Siteng Huang, Jiayan Guo, Xin Li, Yibing Song, Hao Luo, Fan Wang, et~al.
\newblock Worldvla: Towards autoregressive action world model.
\newblock \emph{arXiv preprint arXiv:2506.21539}, 2025.

\bibitem[Chen et~al.(2025{\natexlab{a}})Chen, Su, and Wang]{chen2025wavecl}
Ran Chen, Taiyi Su, and Hanli Wang.
\newblock Wavecl: Wavelet calibration learning for referring video object segmentation.
\newblock In \emph{Proceedings of the 33rd ACM International Conference on Multimedia}, pages 3856--3864, 2025{\natexlab{a}}.

\bibitem[Chen et~al.(2025{\natexlab{b}})Chen, Chen, Chen, Cai, Liu, Li, Liang, Lin, Ge, Gu, et~al.]{chen2025robotwin}
Tianxing Chen, Zanxin Chen, Baijun Chen, Zijian Cai, Yibin Liu, Zixuan Li, Qiwei Liang, Xianliang Lin, Yiheng Ge, Zhenyu Gu, et~al.
\newblock Robotwin 2.0: A scalable data generator and benchmark with strong domain randomization for robust bimanual robotic manipulation.
\newblock \emph{arXiv preprint arXiv:2506.18088}, 2025{\natexlab{b}}.

\bibitem[Chi et~al.(2025)Chi, Jia, Fan, Ju, Mi, Zhang, Qin, Tian, Ge, Li, et~al.]{chi2025wow}
Xiaowei Chi, Peidong Jia, Chun-Kai Fan, Xiaozhu Ju, Weishi Mi, Kevin Zhang, Zhiyuan Qin, Wanxin Tian, Kuangzhi Ge, Hao Li, et~al.
\newblock Wow: Towards a world omniscient world model through embodied interaction.
\newblock \emph{arXiv preprint arXiv:2509.22642}, 2025.

\bibitem[Chung et~al.(2023)Chung, Constant, Garcia, Roberts, Tay, Narang, and Firat]{chung2023unimax}
Hyung~Won Chung, Noah Constant, Xavier Garcia, Adam Roberts, Yi Tay, Sharan Narang, and Orhan Firat.
\newblock Unimax: Fairer and more effective language sampling for large-scale multilingual pretraining.
\newblock \emph{arXiv preprint arXiv:2304.09151}, 2023.

\bibitem[Cuttano et~al.(2025)Cuttano, Trivigno, Rosi, Masone, and Averta]{cuttano2025samwise}
Claudia Cuttano, Gabriele Trivigno, Gabriele Rosi, Carlo Masone, and Giuseppe Averta.
\newblock Samwise: Infusing wisdom in sam2 for text-driven video segmentation.
\newblock In \emph{Proceedings of the Computer Vision and Pattern Recognition Conference}, pages 3395--3405, 2025.

\bibitem[Du et~al.(2023)Du, Yang, Dai, Dai, Nachum, Tenenbaum, Schuurmans, and Abbeel]{du2023learning}
Yilun Du, Sherry Yang, Bo Dai, Hanjun Dai, Ofir Nachum, Josh Tenenbaum, Dale Schuurmans, and Pieter Abbeel.
\newblock Learning universal policies via text-guided video generation.
\newblock \emph{Advances in neural information processing systems}, 36:\penalty0 9156--9172, 2023.

\bibitem[Feng et~al.(2025)Feng, Li, Zhang, Zheng, Luo, Yue, and Lu]{feng2025videoorion}
Yicheng Feng, Yijiang Li, Wanpeng Zhang, Sipeng Zheng, Hao Luo, Zihao Yue, and Zongqing Lu.
\newblock Videoorion: Tokenizing object dynamics in videos.
\newblock In \emph{Proceedings of the IEEE/CVF International Conference on Computer Vision}, pages 20401--20412, 2025.

\bibitem[Gao et~al.(2024)Gao, Yang, Chen, Chitta, Qiu, Geiger, Zhang, and Li]{gao2024vista}
Shenyuan Gao, Jiazhi Yang, Li Chen, Kashyap Chitta, Yihang Qiu, Andreas Geiger, Jun Zhang, and Hongyang Li.
\newblock Vista: A generalizable driving world model with high fidelity and versatile controllability.
\newblock \emph{Advances in Neural Information Processing Systems}, 37:\penalty0 91560--91596, 2024.

\bibitem[Guo et~al.(2025{\natexlab{a}})Guo, Wu, Zhu, Leng, Shi, Chen, Fan, Wang, Jiang, Wang, et~al.]{guo2025seed1}
Dong Guo, Faming Wu, Feida Zhu, Fuxing Leng, Guang Shi, Haobin Chen, Haoqi Fan, Jian Wang, Jianyu Jiang, Jiawei Wang, et~al.
\newblock Seed1. 5-vl technical report.
\newblock \emph{arXiv preprint arXiv:2505.07062}, 2025{\natexlab{a}}.

\bibitem[Guo et~al.(2024)Guo, Hu, Zhang, Wang, Chen, Lu, and Chen]{guo2024prediction}
Yanjiang Guo, Yucheng Hu, Jianke Zhang, Yen-Jen Wang, Xiaoyu Chen, Chaochao Lu, and Jianyu Chen.
\newblock Prediction with action: Visual policy learning via joint denoising process.
\newblock \emph{Advances in Neural Information Processing Systems}, 37:\penalty0 112386--112410, 2024.

\bibitem[Guo et~al.(2025{\natexlab{b}})Guo, Shi, Chen, and Finn]{guo2025ctrl}
Yanjiang Guo, Lucy~Xiaoyang Shi, Jianyu Chen, and Chelsea Finn.
\newblock Ctrl-world: A controllable generative world model for robot manipulation.
\newblock \emph{arXiv preprint arXiv:2510.10125}, 2025{\natexlab{b}}.

\bibitem[Hafner et~al.(2025)Hafner, Pasukonis, Ba, and Lillicrap]{hafner2025mastering}
Danijar Hafner, Jurgis Pasukonis, Jimmy Ba, and Timothy Lillicrap.
\newblock Mastering diverse control tasks through world models.
\newblock \emph{Nature}, pages 1--7, 2025.

\bibitem[Heusel et~al.(2017)Heusel, Ramsauer, Unterthiner, Nessler, and Hochreiter]{heusel2017gans}
Martin Heusel, Hubert Ramsauer, Thomas Unterthiner, Bernhard Nessler, and Sepp Hochreiter.
\newblock Gans trained by a two time-scale update rule converge to a local nash equilibrium.
\newblock \emph{Advances in neural information processing systems}, 30, 2017.

\bibitem[Hu et~al.(2022)Hu, Shen, Wallis, Allen-Zhu, Li, Wang, Wang, Chen, et~al.]{hu2022lora}
Edward~J Hu, Yelong Shen, Phillip Wallis, Zeyuan Allen-Zhu, Yuanzhi Li, Shean Wang, Lu Wang, Weizhu Chen, et~al.
\newblock Lora: Low-rank adaptation of large language models.
\newblock \emph{ICLR}, 1\penalty0 (2):\penalty0 3, 2022.

\bibitem[Huang et~al.(2025)Huang, Chen, Zhou, Chen, Jiang, Hu, Liao, Gao, Li, Yao, et~al.]{huang2025enerverse}
Siyuan Huang, Liliang Chen, Pengfei Zhou, Shengcong Chen, Zhengkai Jiang, Yue Hu, Yue Liao, Peng Gao, Hongsheng Li, Maoqing Yao, et~al.
\newblock Enerverse: Envisioning embodied future space for robotics manipulation.
\newblock \emph{arXiv preprint arXiv:2501.01895}, 2025.

\bibitem[Jiang et~al.(2025)Jiang, Chen, Huang, Chen, Zhou, Liao, He, Liu, Li, Yao, et~al.]{jiang2025enerverse}
Yuxin Jiang, Shengcong Chen, Siyuan Huang, Liliang Chen, Pengfei Zhou, Yue Liao, Xindong He, Chiming Liu, Hongsheng Li, Maoqing Yao, et~al.
\newblock Enerverse-ac: Envisioning embodied environments with action condition.
\newblock \emph{arXiv preprint arXiv:2505.09723}, 2025.

\bibitem[Khoreva et~al.(2018)Khoreva, Rohrbach, and Schiele]{khoreva2018video}
Anna Khoreva, Anna Rohrbach, and Bernt Schiele.
\newblock Video object segmentation with language referring expressions.
\newblock In \emph{Asian conference on computer vision}, pages 123--141. Springer, 2018.

\bibitem[Li et~al.(2025{\natexlab{a}})Li, Gao, Sadigh, and Song]{li2025unified}
Shuang Li, Yihuai Gao, Dorsa Sadigh, and Shuran Song.
\newblock Unified video action model.
\newblock \emph{arXiv preprint arXiv:2503.00200}, 2025{\natexlab{a}}.

\bibitem[Li et~al.(2024)Li, Hsu, Gu, Pertsch, Mees, Walke, Fu, Lunawat, Sieh, Kirmani, et~al.]{li2024evaluating}
Xuanlin Li, Kyle Hsu, Jiayuan Gu, Karl Pertsch, Oier Mees, Homer~Rich Walke, Chuyuan Fu, Ishikaa Lunawat, Isabel Sieh, Sean Kirmani, et~al.
\newblock Evaluating real-world robot manipulation policies in simulation.
\newblock \emph{arXiv preprint arXiv:2405.05941}, 2024.

\bibitem[Li et~al.(2025{\natexlab{b}})Li, Zhu, Wen, Shen, and Xu]{li2025worldeval}
Yaxuan Li, Yichen Zhu, Junjie Wen, Chaomin Shen, and Yi Xu.
\newblock Worldeval: World model as real-world robot policies evaluator.
\newblock \emph{arXiv preprint arXiv:2505.19017}, 2025{\natexlab{b}}.

\bibitem[Liang et~al.(2023)Liang, Wang, Zhou, Miao, Luo, and Yang]{liang2023local}
Chen Liang, Wenguan Wang, Tianfei Zhou, Jiaxu Miao, Yawei Luo, and Yi Yang.
\newblock Local-global context aware transformer for language-guided video segmentation.
\newblock \emph{IEEE Transactions on Pattern Analysis and Machine Intelligence}, 45\penalty0 (8):\penalty0 10055--10069, 2023.

\bibitem[Mu et~al.(2025)Mu, Chen, Chen, Peng, Lan, Gao, Liang, Yu, Zou, Xu, et~al.]{mu2025robotwin}
Yao Mu, Tianxing Chen, Zanxin Chen, Shijia Peng, Zhiqian Lan, Zeyu Gao, Zhixuan Liang, Qiaojun Yu, Yude Zou, Mingkun Xu, et~al.
\newblock Robotwin: Dual-arm robot benchmark with generative digital twins.
\newblock In \emph{Proceedings of the Computer Vision and Pattern Recognition Conference}, pages 27649--27660, 2025.

\bibitem[Peebles and Xie(2023)]{peebles2023scalable}
William Peebles and Saining Xie.
\newblock Scalable diffusion models with transformers.
\newblock In \emph{Proceedings of the IEEE/CVF international conference on computer vision}, pages 4195--4205, 2023.

\bibitem[Pertsch et~al.(2025)Pertsch, Stachowicz, Ichter, Driess, Nair, Vuong, Mees, Finn, and Levine]{pertsch2025fast}
Karl Pertsch, Kyle Stachowicz, Brian Ichter, Danny Driess, Suraj Nair, Quan Vuong, Oier Mees, Chelsea Finn, and Sergey Levine.
\newblock Fast: Efficient action tokenization for vision-language-action models.
\newblock \emph{arXiv preprint arXiv:2501.09747}, 2025.

\bibitem[Quevedo et~al.(2025)Quevedo, Liang, and Yang]{quevedo2025evaluating}
Julian Quevedo, Percy Liang, and Sherry Yang.
\newblock Evaluating robot policies in a world model.
\newblock \emph{arXiv preprint arXiv:2506.00613}, 2025.

\bibitem[Radford et~al.(2021)Radford, Kim, Hallacy, Ramesh, Goh, Agarwal, Sastry, Askell, Mishkin, Clark, et~al.]{radford2021learning}
Alec Radford, Jong~Wook Kim, Chris Hallacy, Aditya Ramesh, Gabriel Goh, Sandhini Agarwal, Girish Sastry, Amanda Askell, Pamela Mishkin, Jack Clark, et~al.
\newblock Learning transferable visual models from natural language supervision.
\newblock In \emph{International conference on machine learning}, pages 8748--8763, 2021.

\bibitem[Seo et~al.(2020)Seo, Lee, and Han]{seo2020urvos}
Seonguk Seo, Joon-Young Lee, and Bohyung Han.
\newblock Urvos: Unified referring video object segmentation network with a large-scale benchmark.
\newblock In \emph{European conference on computer vision}, pages 208--223. Springer, 2020.

\bibitem[Shang et~al.(2025)Shang, Zhang, Tang, Jin, Gao, Wu, and Li]{shang2025roboscape}
Yu Shang, Xin Zhang, Yinzhou Tang, Lei Jin, Chen Gao, Wei Wu, and Yong Li.
\newblock Roboscape: Physics-informed embodied world model.
\newblock \emph{arXiv preprint arXiv:2506.23135}, 2025.

\bibitem[Unterthiner et~al.(2018)Unterthiner, Van~Steenkiste, Kurach, Marinier, Michalski, and Gelly]{unterthiner2018towards}
Thomas Unterthiner, Sjoerd Van~Steenkiste, Karol Kurach, Raphael Marinier, Marcin Michalski, and Sylvain Gelly.
\newblock Towards accurate generative models of video: A new metric \& challenges.
\newblock \emph{arXiv preprint arXiv:1812.01717}, 2018.

\bibitem[Wan et~al.(2025)Wan, Wang, Ai, Wen, Mao, Xie, Chen, Yu, Zhao, Yang, et~al.]{wan2025wan}
Team Wan, Ang Wang, Baole Ai, Bin Wen, Chaojie Mao, Chen-Wei Xie, Di Chen, Feiwu Yu, Haiming Zhao, Jianxiao Yang, et~al.
\newblock Wan: Open and advanced large-scale video generative models.
\newblock \emph{arXiv preprint arXiv:2503.20314}, 2025.

\bibitem[Wang et~al.(2025)Wang, Zhao, Liu, and Chen]{wang2025learning}
Lirui Wang, Kevin Zhao, Chaoqi Liu, and Xinlei Chen.
\newblock Learning real-world action-video dynamics with heterogeneous masked autoregression.
\newblock \emph{arXiv preprint arXiv:2502.04296}, 2025.

\bibitem[Wu et~al.(2024)Wu, Yin, Feng, He, Li, Hao, and Long]{wu2024ivideogpt}
Jialong Wu, Shaofeng Yin, Ningya Feng, Xu He, Dong Li, Jianye Hao, and Mingsheng Long.
\newblock ivideogpt: Interactive videogpts are scalable world models.
\newblock \emph{Advances in Neural Information Processing Systems}, 37:\penalty0 68082--68119, 2024.

\bibitem[Wu et~al.(2025{\natexlab{a}})Wu, Yin, Feng, and Long]{wu2025rlvr}
Jialong Wu, Shaofeng Yin, Ningya Feng, and Mingsheng Long.
\newblock Rlvr-world: Training world models with reinforcement learning.
\newblock In \emph{Advances in Neural Information Processing Systems}, 2025{\natexlab{a}}.

\bibitem[Wu et~al.(2025{\natexlab{b}})Wu, Zhu, Liu, Zhao, Zhai, Cao, and Zha]{wu2025improved}
Pingyu Wu, Kai Zhu, Yu Liu, Liming Zhao, Wei Zhai, Yang Cao, and Zheng-Jun Zha.
\newblock Improved video vae for latent video diffusion model.
\newblock In \emph{Proceedings of the Computer Vision and Pattern Recognition Conference}, pages 18124--18133, 2025{\natexlab{b}}.

\bibitem[Xiang et~al.(2024)Xiang, Liu, Gu, Gao, Ning, Zha, Feng, Tao, Hao, Shi, et~al.]{xiang2024pandora}
Jiannan Xiang, Guangyi Liu, Yi Gu, Qiyue Gao, Yuting Ning, Yuheng Zha, Zeyu Feng, Tianhua Tao, Shibo Hao, Yemin Shi, et~al.
\newblock Pandora: Towards general world model with natural language actions and video states.
\newblock \emph{arXiv preprint arXiv:2406.09455}, 2024.

\bibitem[Yang et~al.(2024)Yang, Teng, Zheng, Ding, Huang, Xu, Yang, Hong, Zhang, Feng, et~al.]{yang2024cogvideox}
Zhuoyi Yang, Jiayan Teng, Wendi Zheng, Ming Ding, Shiyu Huang, Jiazheng Xu, Yuanming Yang, Wenyi Hong, Xiaohan Zhang, Guanyu Feng, et~al.
\newblock Cogvideox: Text-to-video diffusion models with an expert transformer.
\newblock \emph{arXiv preprint arXiv:2408.06072}, 2024.

\bibitem[Yuan et~al.(2024)Yuan, Shi, Yue, and Chen]{yuan2024losh}
Linfeng Yuan, Miaojing Shi, Zijie Yue, and Qijun Chen.
\newblock Losh: Long-short text joint prediction network for referring video object segmentation.
\newblock In \emph{Proceedings of the IEEE/CVF Conference on Computer Vision and Pattern Recognition}, pages 14001--14010, 2024.

\bibitem[Zhang et~al.(2025)Zhang, Peng, Wang, Wang, Zhu, Kang, Jiang, Gao, Li, Liu, et~al.]{zhang2025matrix}
Yifan Zhang, Chunli Peng, Boyang Wang, Puyi Wang, Qingcheng Zhu, Fei Kang, Biao Jiang, Zedong Gao, Eric Li, Yang Liu, et~al.
\newblock Matrix-game: Interactive world foundation model.
\newblock \emph{arXiv preprint arXiv:2506.18701}, 2025.

\bibitem[Zhen et~al.(2025)Zhen, Sun, Zhang, Li, Zhou, Du, and Gan]{zhen2025tesseract}
Haoyu Zhen, Qiao Sun, Hongxin Zhang, Junyan Li, Siyuan Zhou, Yilun Du, and Chuang Gan.
\newblock Tesseract: learning 4d embodied world models.
\newblock \emph{arXiv preprint arXiv:2504.20995}, 2025.

\bibitem[Zheng et~al.(2025)Zheng, Wang, Reed, Bjorck, Fang, Hu, Jang, Kundalia, Lin, Magne, et~al.]{zheng2025flare}
Ruijie Zheng, Jing Wang, Scott Reed, Johan Bjorck, Yu Fang, Fengyuan Hu, Joel Jang, Kaushil Kundalia, Zongyu Lin, Loic Magne, et~al.
\newblock Flare: Robot learning with implicit world modeling.
\newblock \emph{arXiv preprint arXiv:2505.15659}, 2025.

\bibitem[Zheng et~al.(2024)Zheng, Peng, Yang, Shen, Li, Liu, Zhou, Li, and You]{zheng2024open}
Zangwei Zheng, Xiangyu Peng, Tianji Yang, Chenhui Shen, Shenggui Li, Hongxin Liu, Yukun Zhou, Tianyi Li, and Yang You.
\newblock Open-sora: Democratizing efficient video production for all.
\newblock \emph{arXiv preprint arXiv:2412.20404}, 2024.

\bibitem[Zhou et~al.(2024)Zhou, Du, Chen, Li, Yeung, and Gan]{zhou2024robodreamer}
Siyuan Zhou, Yilun Du, Jiaben Chen, Yandong Li, Dit-Yan Yeung, and Chuang Gan.
\newblock Robodreamer: Learning compositional world models for robot imagination.
\newblock \emph{arXiv preprint arXiv:2404.12377}, 2024.

\bibitem[Zhou et~al.(2025)Zhou, Atreya, Tan, Pertsch, and Levine]{zhou2025autoeval}
Zhiyuan Zhou, Pranav Atreya, You~Liang Tan, Karl Pertsch, and Sergey Levine.
\newblock Autoeval: Autonomous evaluation of generalist robot manipulation policies in the real world.
\newblock \emph{arXiv preprint arXiv:2503.24278}, 2025.

\bibitem[Zhu et~al.(2025{\natexlab{a}})Zhu, Yu, Feng, Burchfiel, Shah, and Gupta]{zhu2025unified}
Chuning Zhu, Raymond Yu, Siyuan Feng, Benjamin Burchfiel, Paarth Shah, and Abhishek Gupta.
\newblock Unified world models: Coupling video and action diffusion for pretraining on large robotic datasets.
\newblock \emph{arXiv preprint arXiv:2504.02792}, 2025{\natexlab{a}}.

\bibitem[Zhu et~al.(2025{\natexlab{b}})Zhu, Jia, Gao, Deng, Li, Zhang, Liu, Jia, and Lang]{zhu2025other}
Jian Zhu, Zhengyu Jia, Tian Gao, Jiaxin Deng, Shidi Li, Lang Zhang, Fu Liu, Peng Jia, and Xianpeng Lang.
\newblock Other vehicle trajectories are also needed: A driving world model unifies ego-other vehicle trajectories in video latent space.
\newblock \emph{arXiv preprint arXiv:2503.09215}, 2025{\natexlab{b}}.

\end{thebibliography}


\end{document}